# Cardiovascular Disease Prediction using Recursive Feature Elimination and Gradient Boosting Classification Techniques

Prasannavenkatesan Theerthagiri[*], Vidya J

Department of Computer Science and Engineering, GITAM School of Technology, GITAM University Bengaluru, India. [*]Email: vprasann@gitam.edu

**Abstract:**

Cardiovascular diseases (CVDs) are one of the most common chronic illnesses that affect people's health. Early detection of CVDs can reduce mortality rates by preventing or reducing the severity of the disease. Machine learning algorithms are a promising method for identifying risk factors. This paper proposes a proposed recursive feature elimination-based gradient boosting (RFE-GB) algorithm in order to obtain accurate heart disease prediction. The patients' health record with important CVD features has been analyzed for the evaluation of the results. Several other machine learning methods were also used to build the prediction model, and the results were compared with the proposed model. The results of this proposed model infer that the combined recursive feature elimination and gradient boosting algorithm achieves the highest accuracy (89.7 %). Further, with an area under the curve of 0.84, the proposed RFE-GB algorithm was found superior and had obtained a substantial gain over other techniques. Thus, the proposed RFE-GB algorithm will serve as a prominent model for CVD estimation and treatment.

## 1. Introduction

As the effects of societal aging worsen, health monitoring has become increasingly important. The disease prediction framework can help medicinal experts in anticipating heat alignment in view of the clinical information of patients. Subsequently, by implementing a prediction framework utilizing advanced algorithms and investigating different health-related issues, it can have the capacity to predict more probabilistically that the patients will be diagnosed with any health problems [1].

One of the most significant parts of healthcare monitoring is heart monitoring. A heart disease prediction system can provide valuable insights to medical professionals in making decisions about the state of heart of patients. Aberrant cardiac rhythms can be caused by an abnormal site of origin or irregular conduction of the electric signal. Arrhythmias are the medical term for these diseases. Some arrhythmias can result in significant consequences and even death [1]. Medical professionals may neglect to take exact decision while diagnosing a patient's heart alignment; in this way, a heart alignment prediction method which utilizes machine learning algorithms aid such cases to get precise outcomes [1, 2, 3].

Disease prediction, disease categorization, and medical image recognition algorithms are all examples of machine learning techniques that have been widely applied in medicine [4, 5, 6]. Gradient boosting (GB), a contemporary and efficient method, is presented and enhanced in this paper. The gradient boosting decision tree is the source of the gradient boosting learning method. GB performs well as an ensemble classifier in terms of generalization. Furthermore, GB provides a regularisation term to regulate the model's complexity, which prevents overfitting. In several machine learning disciplines, GB has outperformed the competition [1, 7]. As a result, the performance of GB in classifying single cardiac diseases is investigated.

The goal of this research is to create a clinically useful categorization system for cardiac disease. This work presents a hierarchical technique based on the weighted gradient boosting algorithm to achieve this goal. Preprocessing is the most common method for obtaining useable cardiovascular patient datasets. Following that, numerous types of characteristics are extracted. Following that, recursive feature elimination is used to choose features. Finally, the feature vectors are fed into a hierarchical classifier, which produces predicted labels. The medical field is an application field of information mining because it has a large number of information assets. They realize that it is valuable to include selection and feature reduction. Feature determination is concerned with distinguishing some relevant features sufficient to learn objective thoughts [8, 9, 10].

To choose features for hyperparameter optimization, a stochastic gradient boosting approach is used. To reduce the mean square error, the features are clustered together. Multiple experimental scenarios are examined, and the findings are compared to several earlier studies and typical ML algorithms to prove the usefulness of the suggested technique.

The suggested technique is unique in that it uses the gradient boosting technique to classify cardiovascular diseases. Although the gradient boosting algorithm is well-known, it is processed with the weights of each feature of the dataset for heart disease prediction and classification. The proposed approach is unique in that it uses a hierarchical classifier and recursive feature elimination to choose the best feature from all other features.

The rest of this paper is laid out as follows. The next part provides background information on past efforts as well as an analysis of their shortcomings. With preprocessing, feature selection, and the hierarchical classification approach, Section 3 outlines the proposed approach employed in this study. In Section 4, performance measurements are used to assess the proposed approaches. This section explains the findings and draws some parallels with past research. Section 5 concludes by summarising all of the works and drawing conclusions.

**2. Background**

Artificial intelligence and deep learning algorithms are extremely beneficial for using massive data to predict individual outcomes, especially when coupled to EHRs. This study [1] used machine learning to increase the prediction accuracy of traditional CVD risk variables in a large UK population. The effectiveness of machine learning techniques on longitudinal EHR data for ten-year Cardiovascular event prediction was compared to a gold standard reached through pooled cohort risk [11].

A classification approach with three basic steps was developed in this study [12]. The wavelet approach is used to filter the ECG signal during the preprocessing phase. Then fiducial points are used to find all heartbeats. Feature engineering is a technique for extracting different types of features from time and time-frequency domains. Then, to choose features, this study used recursive feature elimination. To get the final findings, a hierarchical classifier based on the XGBoost classifier and threshold is used in the classification step [12].

The authors devised a prediction approach based on physical examination markers to categorize hypertension patients [13]. The important elements from the patients' many clinical assessment signs are retrieved in the first stage. The essential features retrieved in the first stage are used in the second stage to forecast the patients' outcomes. The authors then suggested a model that incorporated recursive feature removal, cross-validation, and a prediction model. Extreme gradient boosting (XGBoost) is believed to successfully forecast patient outcomes by employing their best features subset [13].

This work [14] proposed a wrapper gene selection strategy with a recursive feature removal approach for efficient classification. For several gene selection strategies, the ensemble technique was used, and the top-ranking genes in each methodology were chosen as the final gene subset. Multiple gene selection techniques were combined in this study, and the ideal gene subset was obtained by prioritizing and ranking the most essential genes picked by the gene selection approach. Consequently, the scientists concluded that selecting a more discriminative and compact gene subset yielded the best results [14].

The scientists used machine learning algorithms to forecast a patient's stage of cardiac disease [15]. They chose the optimal features using the stochastic gradient boosting technique and Recursive Feature Elimination (RFE). An ensemble of weak prediction models, often using decision trees, was used to create a calculation model. It provides a stage-by-stage approach to boost and simplifying, and optimizing a subjectively variational failure problem.

The authors of [16] presented an AutoML approach for automating the process of developing an AI model that performs well on any dataset. This study for cardiovascular disease prediction automates data pre-processing, feature extraction, hyper-parameter tweaking, and algorithm selection. The authors claimed that their AutoML model had removed a significant technical hurdle, allowing doctors to employ AI approaches more widely.

For the best feature detection of the Single Proton Emission Computed Tomography (SPECT), Statlog Heart Disease (STATLOG) datasets, recursive feature removal with cross-validation and stability selection were utilized, and their results were compared [17]. The approaches of Recursive Feature Elimination with Cross-Validation (RFECV) and Stability Selection (SS) were used to enhance the productivity of tree-based and probability-based machine learning techniques in this research. The feature with the lowest score is therefore removed. The RFECV adapts to the RFE and adjusts the number of characteristics picked automatically. The SS method returns details about the output variable's properties. This technique, according to the authors, is most useful in determining the treatment strategy for professionals in the area [17].

To estimate response variables more correctly, the gradient boosting approach fits new models sequentially during learning. The primary concept behind this technique is to build new base-learners with the highest correlation with the ensemble's negative gradient of the error function [18]. Breiman [19] invented Random Forest, an ensemble learning system based on random decision trees. The main distinction between RF and decision trees is that when breaking a node, RF looks for the best feature among the random subsets of characteristics, whereas decision trees look for the most significant feature. As a result, there is a lot of variety, which leads to a better model. The Bayes' theorem-based NB classifier was utilized, with each pair of classified characteristics being independent of one another. In order to discover the most probable categories, it employs probability theory. When the input has high dimensionality, this approach is appropriate [20].

For the Cleveland and Statlog project heart datasets, the authors suggested a model to predict heart disease categorization based on feature selection [21]. According to them, the random forest algorithm's accuracy is good for feature selection (8 and 6 features) based on classification models. Sensitivities and specificity were also associated with higher scores in this study [21].

The gradient boosting decision tree was used in [22] to estimate blood pressure rates based on human physiologic data obtained by the EIMO device. To pick ideal parameters and avoid overfitting, they employed the cross-validation approach. Also, it has been suggested that when considering the features of age, body fat, ratio, and height, this method had displayed with higher accuracy rate with reduced error rates as compared to other algorithms [22]. In this work [23] had proposed a framework for the prediction of risk factors of heart disease using several classifier algorithms. They have revealed that the support vector machine performs with better prediction accuracy, precision, sensitivity, and F1 score [23].

## 3. Proposed Feature Extraction and Classification Technique

This section describes the proposed recursive feature elimination, gradient boosting based machine learning classification technique, feature ranking, and classification/prediction metrics which are used to evaluate the performance of the proposed model.

### 3.1 Dataset of Cardiovascular Disease

The performance of the RFE-GB algorithm is analyzed using the cardiovascular disease dataset retrieved from the Kaggle repository [24]. The cardiovascular disease dataset consists of seventy thousand patient data records with eleven features and a target classifying CVD or non-CVD patients. The eleven attributes are gender, age, height, weight, BP-Systolic, BP-Diastolic, glucose, cholesterol, smoking behavior, physical activities, and patients' alcohol intake. Table.1 summarizes the sample CVD dataset features and their values.

Table.1 A Sample of Cardiovascular Disease Patient Dataset

| Patient/Features | Patient 1 | Patient 2 | Patient 3 | Patient 4 | Patient 5 |
|---|---|---|---|---|---|
| **Age** | 50 | 55 | 52 | 48 | 48 |
| **Gender** | 2 | 1 | 1 | 2 | 1 |
| **Height** | 168 | 156 | 165 | 169 | 156 |
| **Weight** | 62 | 85 | 64 | 82 | 56 |
| **BP (S)** | 110 | 140 | 130 | 150 | 100 |
| **BP (D)** | 80 | 90 | 70 | 100 | 60 |
| **Cholesterol** | 1 | 3 | 3 | 1 | 1 |
| **Glucose** | 1 | 1 | 1 | 1 | 1 |
| **Smoke** | 0 | 0 | 0 | 0 | 0 |
| **Alcohol** | 0 | 0 | 0 | 0 | 0 |
| **Activity** | 1 | 1 | 0 | 1 | 0 |
| **Cardio** | 0 | 1 | 1 | 1 | 0 |

## 3.2 Proposed Methods

The recursive feature elimination algorithm effectively selects the features from the training dataset that are most relevant in target variable prediction. It is an effective method for removing features from a training dataset in preparation for feature selection. RFE is prominent because it is good at identifying the features in a training dataset that are more or less important in predicting the target variable. RFE is a wrapper-style feature selection algorithm that internally employs filter-based feature selection. RFE operates by looking for a subset of features in the training dataset, beginning with all features and successively deleting them until the target number of features [25, 26, 27, 28].

This work builds a model with the predictors, and an importance score is being computed for each predictor. The predictors with a minor significance are removed. Then, the model is rebuilt, and the score is computed again. Here, the number of predictor subsets and their size are specified to evaluate a tuning parameter. The optimal subset can be used to train the model. Thus, the RFE algorithm resulted in the group of top-ranked features that can be considered for selecting features [29]. The dataset has been tested with several subsets of features. It selects the popular features from cardiovascular disease dataset to classify cardio and non-cardio patients with reduced errors.

Figure.1 illustrates the workflow of the proposed RFE-GB methodology for classifying the CVD and Non-CVD patients. The cardiovascular disease dataset has been pre-processed, where missing values are replaced with the mode values. Then normalization was carried out as the dataset contains various measuring units. After normalizing the data, all the attributes of the original data will be in the same order of magnitude. To focus on the relevant data that helps in analysis and prediction, features are selected. This feature elimination strategy supports in a cumulative increase of classification accuracy.

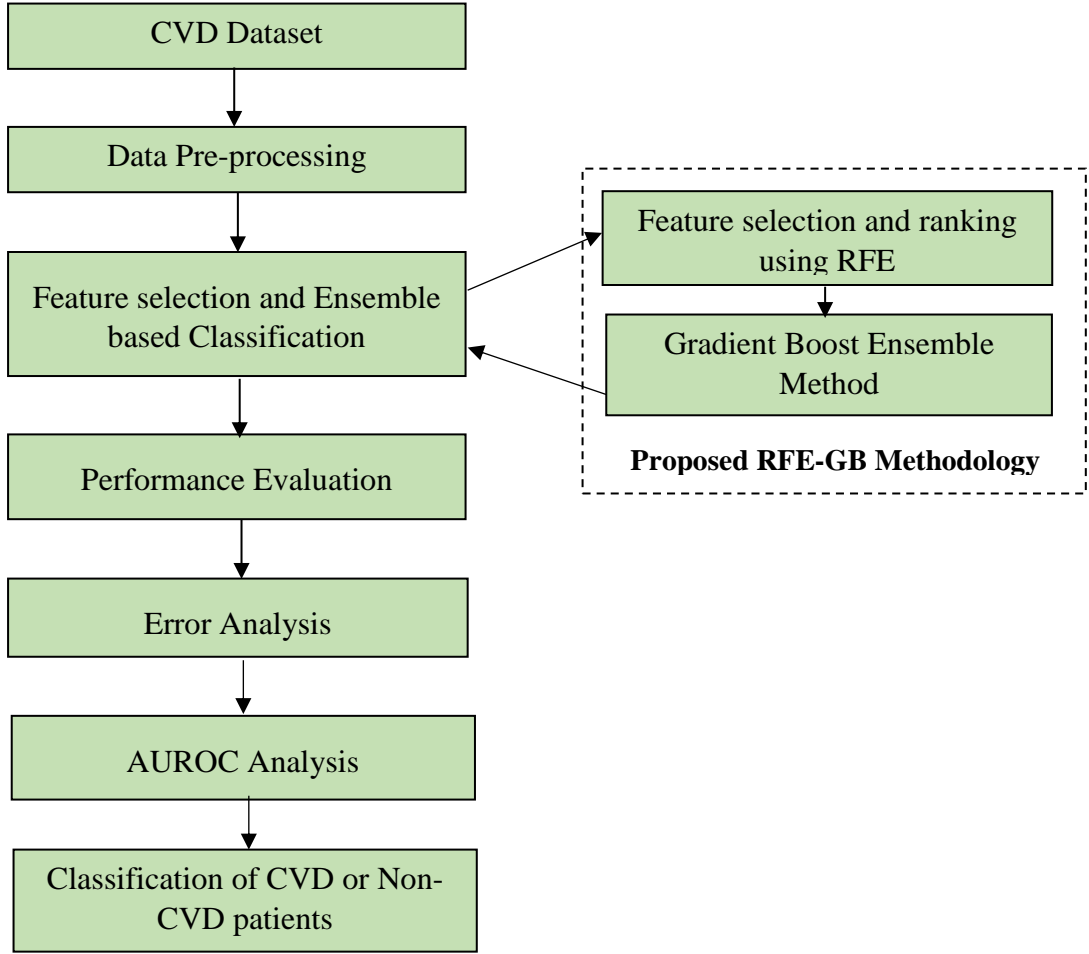

**Figure 1. Workflow of the proposed RFE-GB methodology for CVD classification**

In order to rank the features, the ranking criterion as a separating hyperplane has been determined with the largest margin. Then, a set of training samples are considered, the decision function is given in Equation (1).

$$f(x) = w.x + b \qquad (1)$$

where w is the weight vector which can be obtained by using Equation (2)

$$w = \sum_{i=1}^{n} \alpha_i y_i x_i \qquad (2)$$

where $\alpha_i$ are lagrange multipliers, $x_i \in R^d$ and $y_i \in \{-1,1\}$ and i=1,….n.

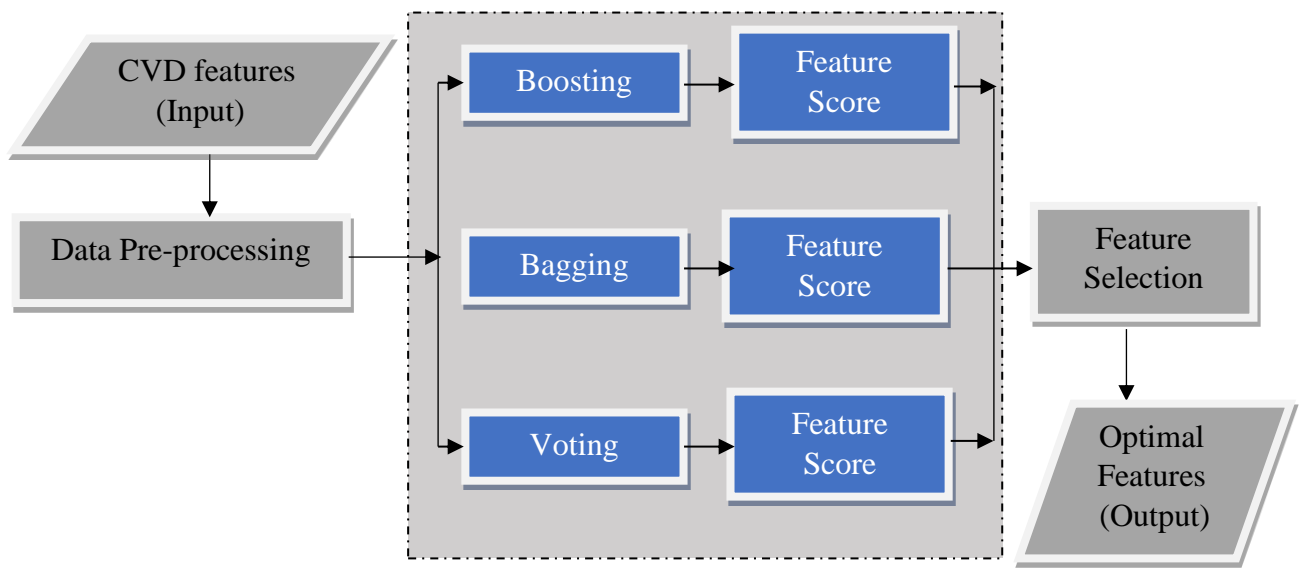

**Fig.2 Feature Selection and Ranking**

The square of the element k of w is used as the kth feature's ranking criterion. Each iteration of RFE is used to train the proposed model, with the lowest-ranking feature being removed because it has the least impact on classification. The remaining features are preserved for the succeeding iteration. This procedure is continued until all of the characteristics have been deleted, at which time they are sorted in the order they were removed. As a result, the attributes that are most important will emerge. Eliminating features one by one will take a long time when the dimension is vast. As a result, more than one feature might be eliminated in each

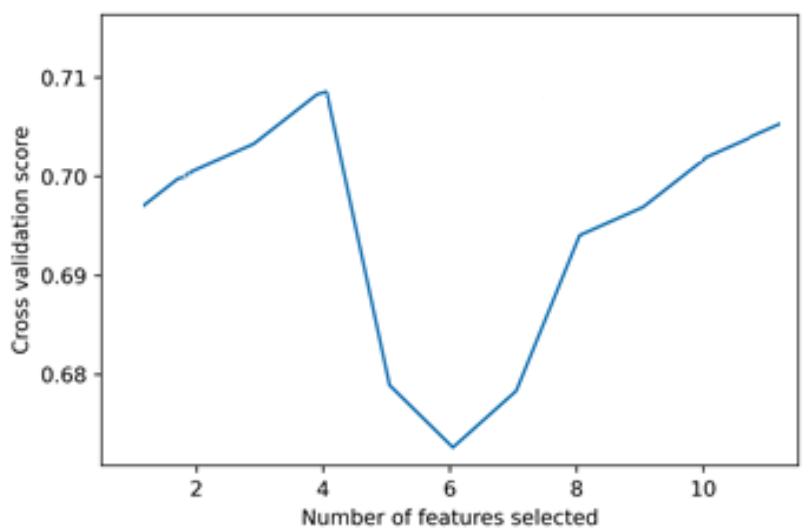

Optimal number of features : 4
Best features : BP(S), BP(D), Cholesterol, Activity

**Fig.3 Feature Selection using RFE-GB algorithm**

iteration, potentially affecting accuracy and causing the correlation bias problem [30]. Figure.2 depicts the feature selection and ranking strategy.

Figure.3 shows the cross-validation scores for the various number of CVD patient features. In this graph, the curve line starts with the cross-validation value of 0.697 and growing up for the three and four features. After reaching the cross-validation value of 0.71, its value starts decreasing for the number of features five, six reaches as lower as 0685. Once, reaching the number of features with six, its cross-validation value again starts to increase. The cross-validation score continues to improve up to eight features. There is a slight deviation, and finally, with twelve features, the curve line touches the cross-validation value nearby 0.71. Thus, this graph clearly depicts the optimal number of features as four with the highest cross-validation score of 0.71. The four best features based on the proposed RFE-GB algorithm are BP(S), BP(D), Cholesterol, and Activity.

Gradient Boosting is the boosting technique based on a decision tree that has been used in this proposed work to solve the classification problem of CVD patients. It sums up weak learners using gradient descent to (it finds the local minimum of the differentiable function) minimize the loss function of the model. As it is an additive model, it generates the learners during the learning process. The impact of a weak learner is determined by the gradient descent optimization procedure. Each tree's contribution is calculated by reducing the strong learner's total error [31]. Consider a gradient boosting technique with X stages and y as the output variable's actual values. Assume an imperfect model Kx at each stage $x$ $(1 \leq x \leq X)$ of gradient boosting. Our method incorporates a new estimator, $h_x(j)$, to enhance Kx as shown in Equations (3), and (4).

$$K_{m+1}(j) = K_m(j) + h_x(j) = y \qquad (3)$$

Thus,

$$h_x(j) = y - K_m(j) \qquad (4)$$

Therefore, gradient boosting will fit h to the residual $y - K_m(j)$ and gives the classification results as to whether CVD patient or non-CVD patient.

**3.3 Performance Evaluation Methods**

In this proposed work, 70% of the data is considered as training data and 30% is taken as testing data from the CVD dataset. To measure the performance of the proposed RFE-GB model, the metrics namely recall, F1-score, precision, confusion matrix, RMSE, AUC-ROC,

Cohen's kappa, and MSE are considered. During error analysis cohorts of data are identified with higher error rates.

The Cohen's Kappa score is an excellent measure to handle multi-class and imbalanced class problems very well. Its value ranges from zero to one, and it is derived using Equation (3), where, $p_o$ is observed class and $p_e$ is the expected class of CVD patients. The Mean Squared Error (MSE) gives the average error between actual and predicted values. Its value can be determined by taking the average square of the difference between the original and predicted values. Equation (4) gives the way to calculate the MSE, where 'n' represents the number of CVD patient records in the dataset. The Root Mean Squared Error (RMSE) gives the square root of the average error between actual and predicted values. Its value can be determined with Equation (5).

$$k = \frac{p_o - p_e}{1 - P_e} = 1 - \frac{1 - p_o}{1 - p_e} \qquad (3)$$

$$MSE = \frac{1}{n}\sum_{i=1}^{n}((actual\ values - predicted\ values)^2) \qquad (4)$$

$$RMSE = \sqrt{\frac{\sum_{i=1}^{n}(predicted_i - actual_i)^2}{n}} \qquad (5)$$

In this work, the Area Under the Curve- Receiver Operator Characteristic (AUC-ROC) measures a classifier's ability to differentiate between CVD and non-CVD classes. The Receiver Operator Characteristic is a probability curve that plots the TPR (True Positive Rate) against FPR (False Positive Rate) at different threshold values. The confusion matrix is an nxn matrix that evaluates the performance of a classification model, and it compares the actual target values with the values predicted by the machine learning model. Further, the precision, recall, and F1-score are analyzed for the proposed RFE-GB model. The precision gives the ratio between the TP (True Positive) and all the positives (True Positive and False Positive) as given in Equation (6). The recall is the measure of how the model correctly detecting True Positives. Its way of calculation is given in Equation (7). The F1-score is the harmonic mean of the recall and precision, and it is presented in Equation (8).

$$Precision = \frac{True\ Positive}{True\ Positive + False\ Positive} \qquad (6)$$

$$Recall = \frac{True\ Positive}{True\ Positive + False\ Negative} \qquad (7)$$

$$F1-score = 2*\frac{Precision*Recall}{Precision+Recall} \qquad (8)$$

**4. Discussions and Findings**

This section presents the cardiovascular disease prediction results of the proposed Recursive Feature Elimination based Gradient Boosting (RFE-GB) classification algorithm with the traditional Linear Discriminant Analysis (LDA), K-Nearest Neighbours (KNN), "Decision Tree (DT), Naive Bayes (NB), and Multilayer perceptron (MLP) algorithms. The resampling method used in this study to verify the machine learning algorithms' research results is k-fold cross-validation. The 'k' value in this work is set to 10. As a result, it's often referred to as a 10-fold cross-validation resampling process. The 10-fold cross-validation approach is designed to reduce the prediction model's bias.

**4.1 Performance Evaluation**

The effectiveness of machine learning prediction algorithms is often measured using a set of classification algorithm-based metrics. The prediction error rates are quantified using the mean square error (MSE), root mean square error (RMSE), and Kappa score in this study. The confusion matrix and receiver operating characteristic area under the curve are used to analyze the predictions' true/false positive/negative rate (ROC AUC). The machine learning algorithms' prediction performance is measured by prediction accuracy, precision, recall, and f1 score [18, 20, 23].

**Table.2 Performance metrics**

| S.No. | Algorithm/metrics | Accuracy | Precision | Recall | F1-score |
|---|---|---|---|---|---|
| 1 | Linear discriminant analysis (LDA) | 64.87 | 0.67 | 0.61 | 0.64 |
| 2 | K-Neighbors Classifier (KNN) | 68.74 | 0.7 | 0.66 | 0.68 |
| 3 | Decision Tree (DT) | 63.58 | 0.65 | 0.62 | 0.63 |
| 4 | Naive Bayes (NB) | 58.3 | 0.74 | 0.27 | 0.4 |
| 5 | Multi-Layer Perceptron (MLP) | 76.42 | 0.72 | 0.72 | 0.72 |
| 6 | **Proposed RFE-GB** | **89.78** | **0.86** | **0.84** | **0.83** |

Importantly, the accuracy of the above-mentioned machine algorithms is determined in this study (whether the patient has cardiovascular disease or not). Each classification model has a distinctive disease prediction accuracy and efficiency over other prediction models based on its hyperparameters. 70% of the dataset is utilized for training, whereas 30% of the data samples are utilized to test classification methods in this study. The proposed disease classification reports of the RFE-GB algorithm are compared with traditional machine learning models, and it is presented in Table.2.

Table.2 summarizes the performance results namely prediction accuracy, precision, recall, and F1-score. The proposed recursive feature elimination-based gradient boosting algorithm has higher accuracy, precision, recall, and F1-score as 89.78, 0.86, 0.84, and 0.83, respectively. In contrast, other machine learning algorithms produce lower results as compared with the proposed RFE-GB Algorithm. Such that the proposed recursive feature elimination based gradient boosting algorithm produces the prediction accuracy of 89.78 %, whereas the other machine learning classification algorithms linear discriminant analysis, k-nearest neighbors, decision tree, Naive Bayes, and multilayer perceptron resulted with 64.87, 68.74, 63.58, 58.3, and 76.42 % of respective prediction accuracy only.

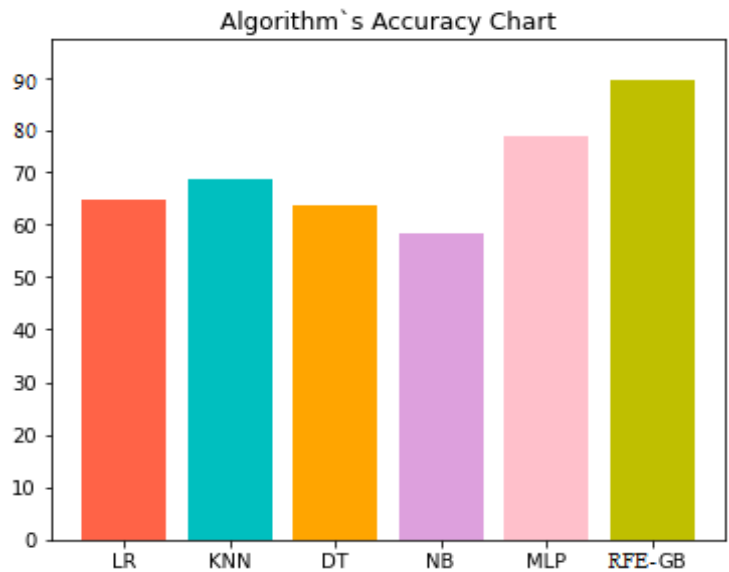

**Fig.4 Prediction Accuracy of ML algorithms**

Among that, the Naive Bayes algorithm performs worst with the accuracy of 58.3 %, precision, recall, F1-score of 0.74, 0.27, 0.4, respectively. Figure.4 gives the accuracy of the proposed RFE-GB and other existing algorithms. We can see that the proposed recursive feature elimination based gradient boosting algorithm predicts the cardiovascular disease patients about 90 % (based on patient's age, gender, height, weight, systolic blood pressure, diastolic blood pressure, cholesterol, glucose, smoke, alcohol intake, and physical activity) more accurately than the other algorithms.

Further, Figure.4 clearly states that the proposed RFE-GB algorithm has the highest accuracy of 89.78. The proposed RFE-GB algorithm operates by looking for a subset of features in the training dataset, beginning with all of the features and successfully deleting them until the target number of features. The cardiovascular disease dataset has been tested with several subsets of features. It selects the popular features from the dataset to classify cardio and non-cardio patients with reduced errors. As a result, the proposed recursive feature elimination-based gradient boosting technique outperforms the other classification rate methods. As a result, the proposed RFE-GB algorithm outperforms the LDA, KNN, DT, NB, and MLP algorithms by 13.36 percent to 31.48 percent. Figure.5 gives the performance of precision and recall. It can be seen that both metrics have the highest scores as 0.84 and 0.83, than others. Further, the F1-score of the proposed RFE-GB algorithm has 11 % to 43 % of improved results.

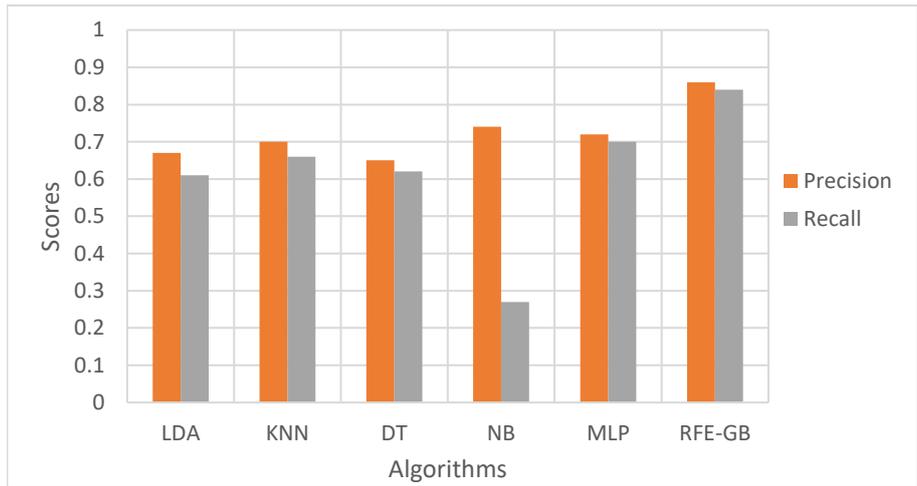

**Fig.5 Classification report**

Cohen's kappa score for the proposed and existing machine learning algorithms depicted in Figure.6; the proposed RFE-GB method has a higher kappa score than traditional methods, as seen by the graph. Cohen's kappa score assures that the classification algorithm's predictions are consistent [32]. Furthermore, it indicates that among the experimented algorithms, the proposed RFE-GB algorithm generates the highest consistency (Kappa score) of 0.57; whereas 0.3973, 0.3125, 0.5112, 0.3770, and 0.3213 are the values for the KNN, Naive Bayes, extra trees, decision trees, and radial base function, respectively.

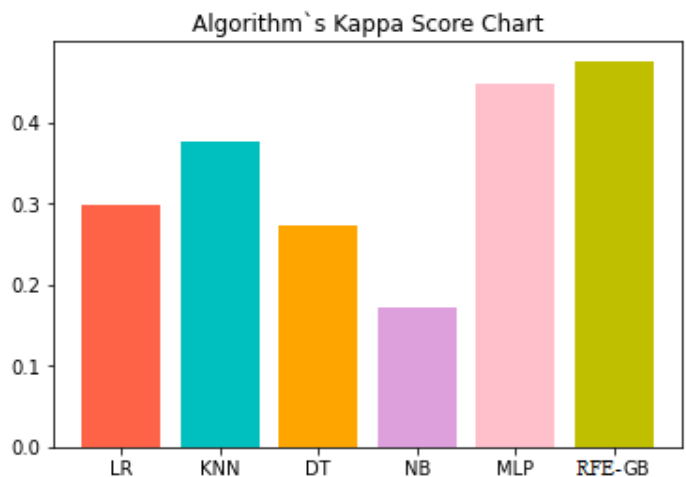

**Fig.6 Cohen's kappa scores**

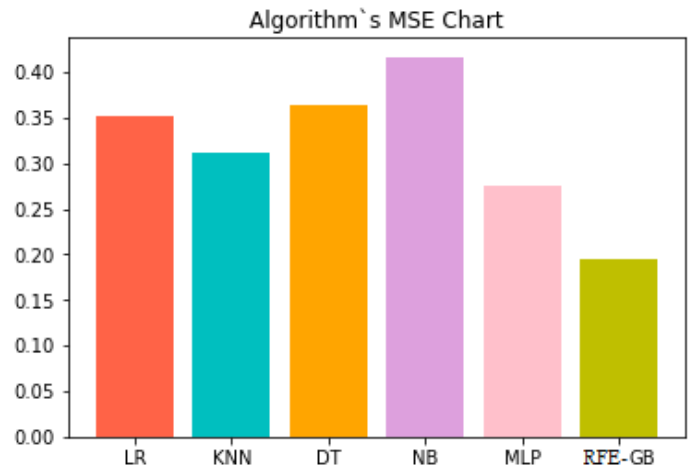

**Fig.7 MSE rates**

Table.3 shows the mean square error (MSE), and root mean square error (RMSE) values for the various machine learning techniques. The MSE error rates of the LDA, KNN, DT, Naive Bayes, MLP, and proposed RFE-GB algorithms are respectively 0.35129, 0.31257, 0.36414, 0.41693, 0.27579, and 0.19243. It has the lowest error rate of 0.1924 for predicting correct cardiac disease instances when compared to the other algorithms (Figure.7). The proposed RFE-GB algorithm boosts the weak features from the higher-ranked and selected features in the training and testing dataset. As a result, error rates are reduced. Likewise, the RMSE error rate of the proposed RFE-GB algorithm also very low (0.43), as illustrated in Table.3; whereas the error rates for other algorithms are LDA (0.59), KNN (0.55), DT (0.60), NB (0.64), and MLP (0.52).

**Table.3 Error performance metrics**

| S.No. | Algorithm/metrics | MSE | RMSE |
|---|---|---|---|
| 1 | Linear discriminant analysis (LDA) | 0.35129 | 0.59269 |
| 2 | K-Neighbors Classifier (KNN) | 0.31257 | 0.55908 |
| 3 | Decision Tree (DT) | 0.36414 | 0.60344 |
| 4 | Naive Bayes (NB) | 0.41693 | 0.6457 |
| 5 | Multi-Layer Perceptron (MLP) | 0.27579 | 0.52515 |
| 6 | **Proposed RFE-GB** | **0.19243** | **0.43866** |

The confusion matrix for various machine learning algorithms is shown in Figure 8. The percentage of predicted values and the percentage of real values are represented in terms of true positives/negatives and false positives/negatives in this graph. The proposed RFE-GB algorithm accurately estimates 88 percent (true positive) of cardio cases, with just 12 percent

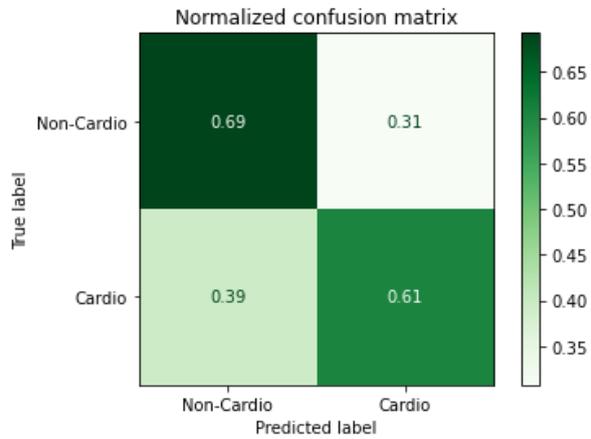
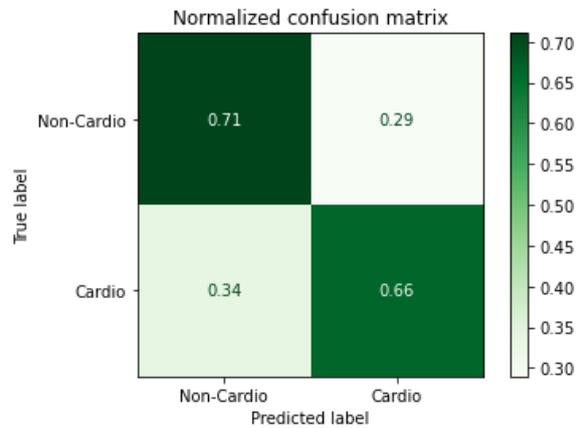

**Fig.8 (a) LDA**                **Fig.8 (b) KNN**

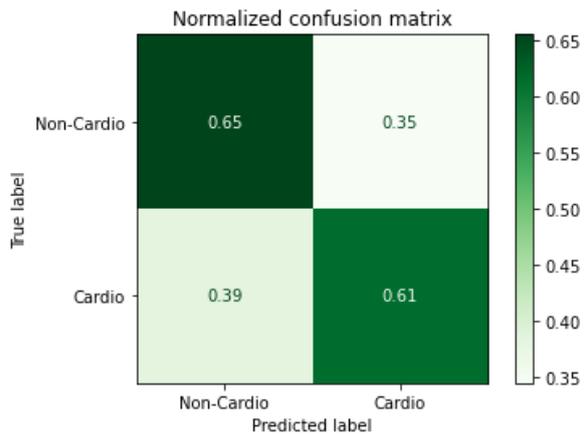
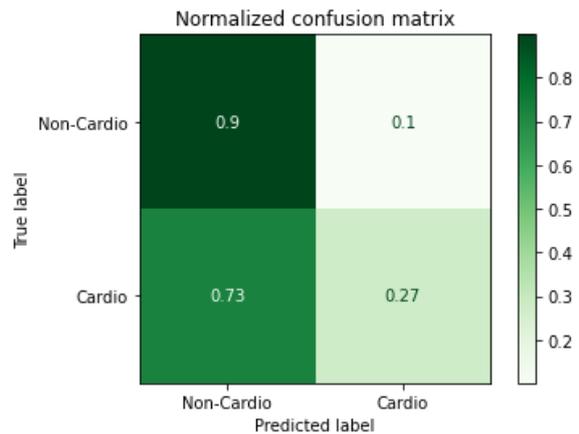

**Fig.8 (c) DT**                **Fig.8 (d) NB**

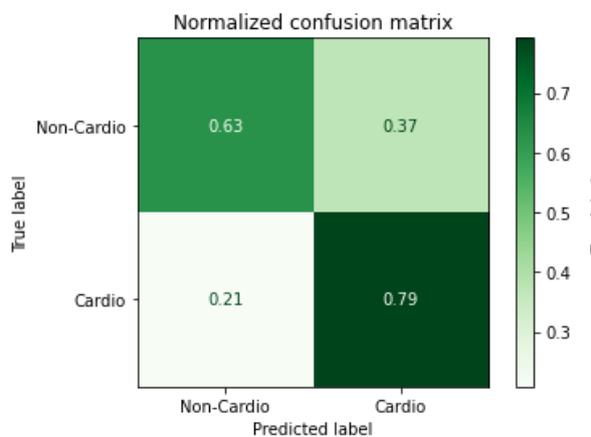
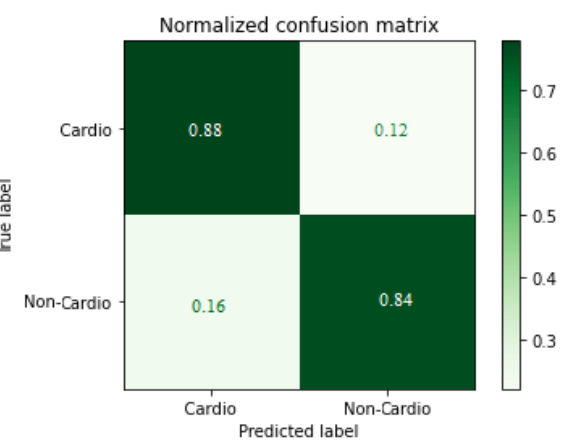

**Fig.8 (e) MLP**                **Fig.8 (f) Proposed RFE-GB**

**Fig.8 Normalized Confusion matrix of ML algorithms**

(false positive) misclassification; for the non-cardio cases, the RFE-GB algorithm gives 16 percent (false negative) of misclassification and 84 percent (true negative) of precise classification as illustrated in Figure.8(f). Similarly, Figure. 8(a), Figure.8(b), Figure.8(c), Figure.8(d), and Figure.8(e) depict the confusion matrix of LDA, KNN, NB, DT, and MLP algorithms respectively, with lower true positives/negatives and false positives/negatives rate than the RFE-GB algorithm.

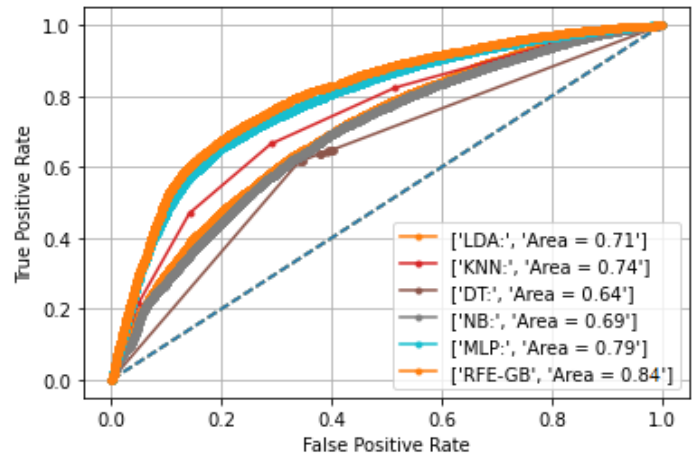

**Fig.9 ROC_AUC Curve**

In the form of a ROC area under the curve, Figure.9 depicts the relationship between the false positive rate and the true positive rate. As opposed to LDA (0.71), KNN (0.74), NB (0.69), DT (0.64), and MLP (0.79) algorithms, the RFE-GB algorithm generates the highest value of 0.84. These results prove that the proposed RFE-GB algorithm accurately classifies cardiovascular disease patients based on their health records.

**5. Conclusion**

It is worth researching much of what is required to forecast and diagnose any disease using machine learning effectively. This work recursive feature elimination-based gradient boosting algorithm has been proposed to select the most important features from the cardiovascular disease dataset. The RFE-GB algorithm selects three optimal number of features as blood pressure, cholesterol, and physical activity from the 12 features. Adopting these three features, a gradient boosting ensemble approach has been developed to predict cardiovascular disease cases. The proposed RFE-GB algorithm has been evaluated with various metrics, and its performance results are compared with explores different machine learning algorithms.

Among that, the proposed RFE-GB algorithm has 13.36 % to 31.48 % of improved accuracy as compared to LDA, KNN, DT, NB, and MLP algorithms.

Further, it produces higher consistency (Kappa score) of 0.57 with a reduced error rate MSE of 0.1924 on the prediction of accurate cardio disease cases. The proposed RFE-GB algorithm accurately estimates 88 percent true positives and 84 percent of true negatives from 70,000 patient records with the AUROC score of 84 %. As a consequence of the findings, the proposed RFE-GB algorithm appears to be capable of diagnosing and classifying diabetes patients. Various combinations of machine learning approaches can be used for the future plans of this research.